\documentclass[journal]{IEEEtran}
\ifCLASSINFOpdf
\else
\fi

\usepackage{todonotes}
\usepackage{subcaption}
\usepackage{graphicx}
\usepackage{color}
\usepackage{cite}
\usepackage{comment}
\usepackage{caption}
\usepackage{makecell}

\usepackage{arydshln}
\usepackage{amsmath}
\usepackage{amssymb}

\usepackage{siunitx}
\begin{document}
\title{Breaking the Limits of Remote Sensing by Simulation and Deep Learning for Flood and Debris Flow Mapping}

\author{Naoto~Yokoya, 
        Kazuki~Yamanoi, 
        Wei~He, 
        Gerald~Baier, 
        Bruno~Adriano, 
        Hiroyuki~Miura, 
        and~Satoru~Oishi% <-this % stops a space
%\thanks{This work was supported in part by the Japan Society for the Promotion of Science through KAKENHI under Grants 18K18067, 19K20308, 19H02408, 20K19834, ...}
\thanks{N. Yokoya is with the Department of Complexity Science and Engineering, Graduate School of Frontier Sciences, the University of Tokyo, Chiba 277-8561, Japan, and also with the RIKEN Center for Advanced Intelligence Project, Tokyo, Japan. (e-mail: yokoya@k.u-tokyo.ac.jp)}% <-this % stops a space
\thanks{Kazuki Yamanoi is with the Disaster Prevention Research Institute, Kyoto University, Kyoto 612-8235, Japan and also a visiting researcher in the RIKEN Center for Computational Science, Kobe, Japan (yamanoi.kazuki.6s@kyoto-u.ac.jp).}% <-this % stops a space
\thanks{Wei He, Gerald Baier, and Bruno Adriano are with the RIKEN Center for Advanced Intelligence Project, Tokyo 103-0027, Japan. (email: \{wei.he,gerald.baier,bruno.adriano\}@riken.jp)}% <-this % stops a space
\thanks{Hiroyuki Miura is with the Graduate School of Advanced Science and Engineering, Hiroshima University, Hiroshima 739-8527, Japan. e-mail: hmiura@hiroshima-u.ac.jp)}
\thanks{Satoru Oishi is with the RIKEN Center for Computational Science, Kobe 650-0047, Japan. (email: satoru.oishi@riken.jp).}% <-this % stops a space
%\thanks{Manuscript received ---}
}


\markboth{}%Journal of \LaTeX\ Class Files,~Vol.~X, No.~X, X~202X}%
{Yokoya \MakeLowercase{\textit{et al.}}: Bare Demo of IEEEtran.cls for IEEE Journals}



\maketitle

\begin{abstract}
We propose a framework that estimates inundation depth (maximum water level) and debris-flow-induced topographic deformation from remote sensing imagery by integrating deep learning and numerical simulation.
A water and debris flow simulator generates training data for various artificial disaster scenarios.
We show that regression models based on Attention U-Net and LinkNet architectures trained on such synthetic data can predict the maximum water level and topographic deformation from a remote sensing-derived change detection map and a digital elevation model.
The proposed framework has an inpainting capability, thus mitigating the false negatives that are inevitable in remote sensing image analysis.
Our framework breaks the limits of remote sensing and enables rapid estimation of inundation depth and topographic deformation, essential information for emergency response, including rescue and relief activities.
 We conduct experiments with both synthetic and real data for two disaster events that caused simultaneous flooding and debris flows and demonstrate the effectiveness of our approach quantitatively and qualitatively.
%\textcolor{red}{The datasets used in this paper are made freely available (after acceptance).}
\end{abstract}


% \fi
\IEEEpeerreviewmaketitle

\section{Introduction}
Information about inundation depth (maximum water level) and debris flow-induced topographic deformation is essential for flood and debris flow emergency response, including rescue and relief activities. Floods and debris flows often occur jointly following torrential rain, and their complexity makes damage assessment challenging~\cite{nhess-18-2161-2018}.
Most remote sensing-based techniques for rapid mapping have been limited to detecting the spatial extent of flood and debris flow~\cite{rs70809822,SCHLAFFER201515,LI2018123,TONG2018144,wieland2019modular,li2019urban}. Numerical simulation models are capable of calculating the realistic maximum water level and topographic deformation; however, they require accurate input data and time-consuming parameter tuning.
%Although it is possible to obtain data on the maximum water level and topographic deformation through field surveys and LiDAR measurements, such data are rare due to their high cost.

\begin{figure}[t]
\centering
\includegraphics[width=\linewidth]{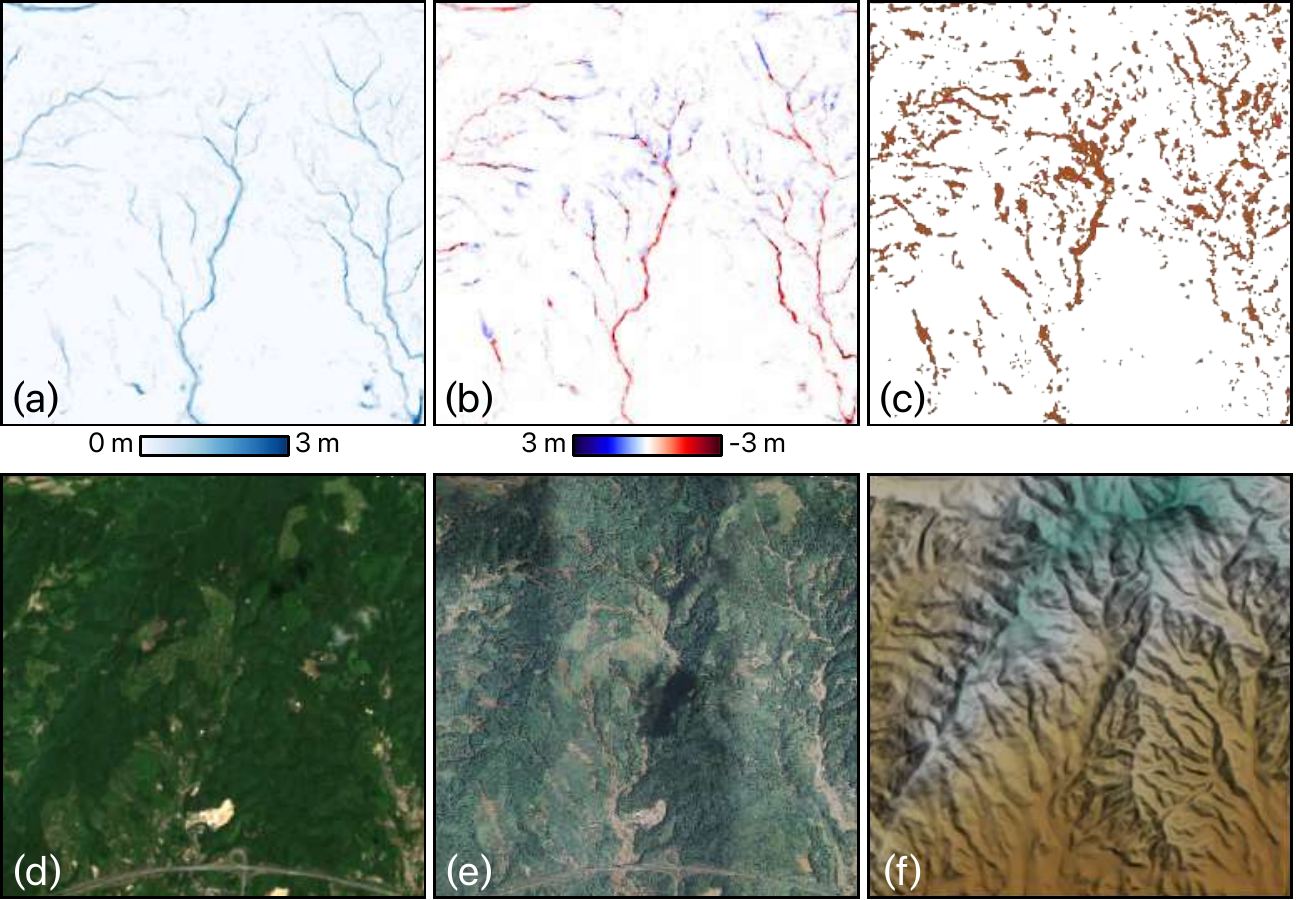}
\caption{Our framework enables to estimate (a) the maximum water level and (b) topographic deformation from (c) the binary change map derived from (d) pre- and (e) post-disaster images together with (f) DEM.}
\label{methodology}
\end{figure}

In this paper, we propose a framework that estimates the maximum water level and topographic deformation after simultaneous rainfall-triggered flood and debris flow, using remote sensing imagery and a combination of numerical simulation and deep learning.
We synthesize training data comprising triplets of maximum water level, topographic deformation, and binary change maps for various artificial disaster scenarios by simulation and binarization of change information. A regression model bridges simulation and observation, inferring the maximum water level and topographic deformation from ground-surface change information derived from remote sensing imagery and topographic data, i.e., a digital elevation model (DEM).
By training this type of inverse estimation model in advance, it is possible to obtain detailed  maximum water level and topographic deformation as soon as remote sensing imagery is available after a disaster.

The advantage of the proposed framework is that it breaks existing limits of remote sensing and enables rapid estimation of detailed disaster information, such as the maximum water level and debris-flow-induced topographic deformation.
Using simulations circumvents the need to obtain real training data, which is expensive and complicated by the very nature of disasters, which are rare occurrences in which affected areas are difficult to access.
By combining deep learning and numerical simulations, the proposed framework also learns characteristics of the analyzed quantities, such as shape and location.
This mitigates the false negatives that are often inevitable in change maps derived by automated remote sensing image analysis.
Our contributions are threefold.
\begin{itemize}
    \item We propose a framework that integrates remote sensing, deep learning, and numerical simulation to estimate the maximum water level and topographic deformation after floods and debris flows.
    \item We construct two datasets for our task based on real data collected after two complex disasters characterized by floods and debris flows, following torrential rains in Japan. %and \textcolor{red}{make these datasets freely available to the community.}
    \item We evaluate our methodology with two real cases and demonstrate its effectiveness both qualitatively and quantitatively.
\end{itemize}

The remainder of this paper is organized as follows. Section II provides an overview of related work. Section III introduces our methodology. Section IV presents the experimental results, and Section V concludes the paper with some remarks and thought about plausible future lines of research.

\section{Related Work}
\subsection{Flood and Landslide Mapping via Remote Sensing}
%=======================
% flood mapping
%=======================
Flood detection by remote sensing has been well studied, and several systems\footnote{https://floodobservatory.colorado.edu/index.html}
\footnote{https://floodmap.modaps.eosdis.nasa.gov/}
\footnote{https://www.dlr.de/eoc/en/desktopdefault.aspx/tabid-12939/22596\_read-51634/} operate on a global scale.
Flood detection at high resolution is mainly based on synthetic aperture radar (SAR) images or optical images, and can be broadly divided into unsupervised and supervised approaches.
In unsupervised approaches~\cite{martinis2009towards,martinis2010unsupervised,sakamoto2007detecting}, pre- and post-disaster images (e.g., intensity images of SAR), index images (e.g., spectral indices of optical data), or their differences are thresholded and then smoothed or masked out to mitigate false positives and false negatives.
Supervised approaches~\cite{TONG2018144,mueller2016water} identify flooded areas by detecting water in the pre- and post-disaster images using pixel-wise classification (or semantic segmentation).
Wieland and Martinis developed a fully automated system based on a convolutional neural network (CNN) to automatically detect floods from multispectral images~\cite{wieland2019modular}. Flood detection in urban areas using SAR images is challenging, and Li et al. tackled this problem with an active self-learning CNN \cite{li2019urban}. On the other hand, Ohki et al. ~\cite{8949490} used a SAR interferometric phase statistics to estimate the flood segments in built-up areas.
Cohen et al. developed a methodology that estimates water level from a flood inundation map and DEM for fluvial floods~\cite{cohen2018estimating}. The estimation of maximum water level from remote sensing images remains a challenging task when there are false negatives in a flood inundation map and also for flash floods due to the dynamics of water.

%=======================
% landslide mapping
%=======================
Detection of landslides, including debris flows, is another common topic in remote sensing image analysis for disaster mapping.
%In machine learning-based landslide susceptibility analysis, remote-sensing images have been frequently used as input data along with topography slope, land cover, rainfall data, and seismic activity~\cite{GLC, 4069113, Kirschbaum2010, Kirschbaum2015, Kirschbaum2018}.
As with flood detection, change detection using pre- and post-disaster images is a typical approach. The use of spectral indices from multispectral images (e.g., normalized vegetation and soil index) is the simplest and effective method of detecting landslides, particularly in vegetated mountainous areas~\cite{6506977, Lv2018, 6341795, 8620663, Ramos-Bernal2018}.
Landslide detection using SAR intensity imagery is an alternative to optical image-based approaches in adverse weather conditions.
But its accuracy is limited by the presence of layovers and shadowing particularly for narrow debris flows~\cite{Shi2015,Darvishi2018, Mondini2019,rs12030561}.
Interferometric SAR has been demonstrated to be advantageous in detecting large-scale, slowly moving landslides~\cite{zhao2012large}.
Research on landslide detection using machine learning from optical and SAR images has recently gained popularity~\cite{Bui2018, Park2018, Burrows2019, Ghorbanzadeh2019, Wang2019}, but the collection of training data is costly.
An effective means of estimating more detailed damage information, such as the amount of soil runoff and deposition, is to analyze the topography before and after the disaster using LiDAR~\cite{miura2019fusion}.
However, LiDAR measurements are costly and thus usually not available from the initial observations of disaster areas by aircraft and helicopters.
Therefore, it remains challenging to estimate debris flow-induced topographic deformation from emergency observations.
In this work, the synergistic use of deep learning and numerical simulation provides a solution to the above problems.

\subsection{Flood and Debris Flow Simulation}
The simulation of flood hazards has been well studied and is already a common technique for estimating flood risk. Traditionally, most simulation methods require inflow to the area of interest as a boundary condition, but some methods have been developed to predict from observable rainfall data by integrating with the rain runoff process\cite{Yamazaki2011,Sayama2012}. However, such methods, which deal only with water, are insufficient to simulate debris flow that consists of water and sediment materials.

Simulation methods for debris flow have also been developed by several research groups. The most typical method tracks the debris flow from a certain inflow point based on the fluid dynamics method \cite{Rickenmann2006, Chen2017, Han2018, Bao2019,Nakatani2016,Frank2015,Gao2016}. In these methods, the location and flow discharge should be given in order for simulations to be conducted; however, these data are normally based on observational information, such as debris flow trace, which can be obtained in post-disaster terms. Therefore, the usability of these simulations is not high enough for prediction purposes.

In contrast, methods that estimate both the transportation and development of debris flow \cite{Revellino2004, Schraml2015, Rodriguez-Morata2019} can be applied only from the initial location of the slope failures. By connecting with a statistical landslide prediction, predictive simulation that requires no debris flow traces has also been proposed \cite{yamanoi2020preprint}. In this work, we employed this predictive method to generate several scenarios of rainfall-triggered flood and debris flow damages.
%the debris flow damage data for training regression models.

\begin{figure*}[t]
\centering
\includegraphics[width=\linewidth]{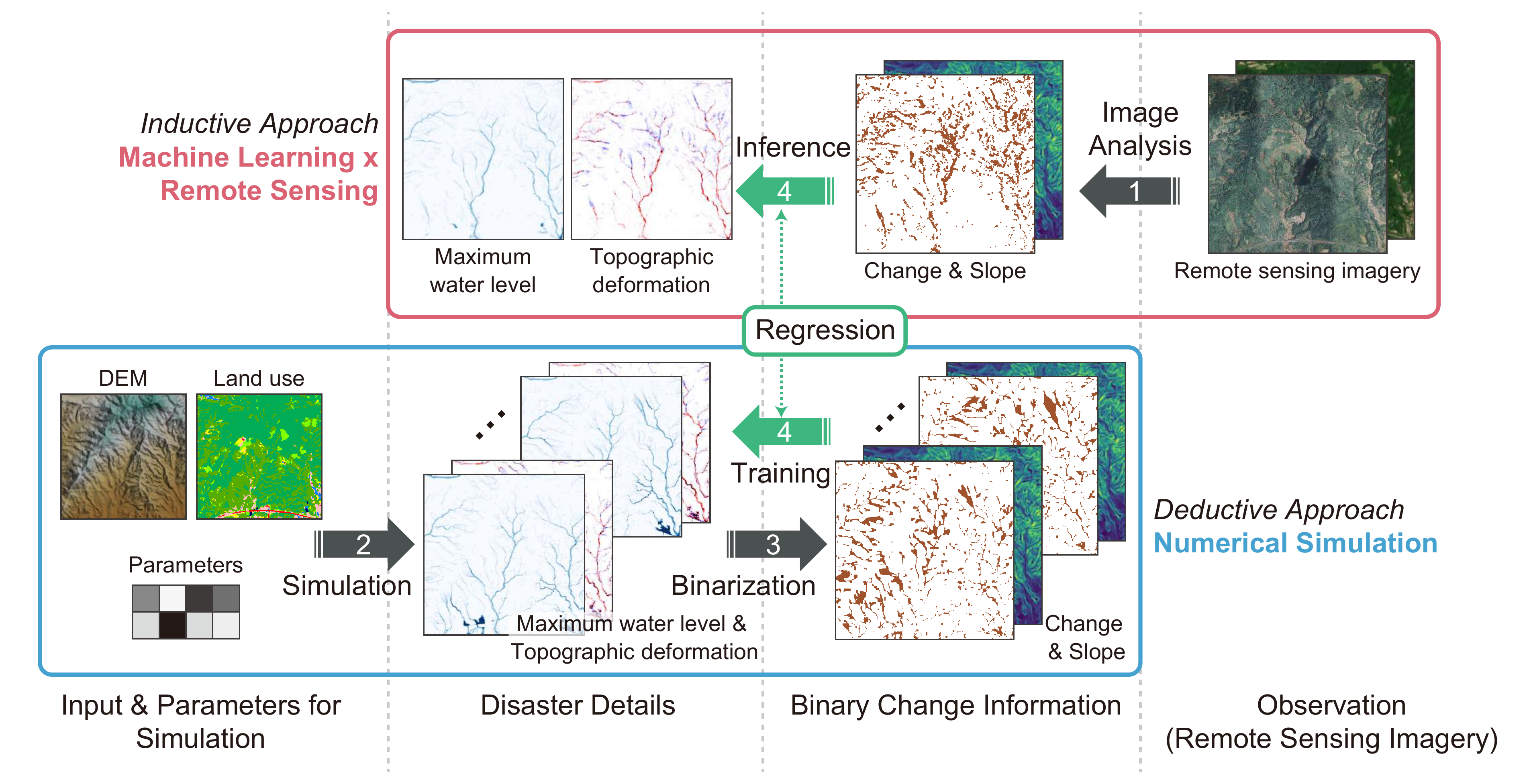}
\caption{The overview concept of the proposed framework.}
\label{methodology}
\end{figure*}

\subsection{Synergy of Deep Learning and Simulation}
The collection of training data is a challenge for deep learning in all fields. Enormous efforts have been made to create synthetic data for training through simulation in various fields, including computer vision, bioinformatics~\cite{ching2018opportunities}, natural language processing~\cite{devlin2018bert,dai2019transformer}, and remote sensing~\cite{anantrasirichai2019deep}. In computer vision, for example, simulation-generated synthetic data are widely used as training data for basic tasks such as depth estimation~\cite{zheng2018t2net}, optical flow~\cite{ilg2017flownet}, semantic segmentation~\cite{tobin2017domain,tsai2018learning}, and object detection~\cite{bak2018domain}. Full-scale simulation environments have been used to create indoor and outdoor scenes for autonomous driving~\cite{li2019aads,hurl2019precise}, robotics~\cite{gao2019vrkitchen,chociej2019orrb}, and aerial navigation~\cite{sadeghi2016cad2rl,krishnan2019air}. Research on domain adaptation is also underway to more efficiently utilize models learned from synthetic data for the analysis of real data~\cite{tobin2017domain,bak2018domain}.

Collecting training data for very rare events such as disasters is challenging. In particular, it is difficult to collect dense, detailed disaster information from real measurements, such as the inundation depth and topographic deformation.
Inspired by the above-mentioned research, this work proposes to generate training data for flood and debris flow by numerical simulation.
In the event of a disaster, a deep model can then rapidly estimate detailed damage information, using a change detection map obtained by conventional remote sensing image analysis as input.

\section{Methodology}
To break the limits of current remote sensing approaches, we combine two technologies: numerical simulation and deep learning.
The former can generate a sufficient amount of synthetic data for training, and the later is capable of solving complex inverse problems from a significant amount of training data. The proposed methodology comprises four modules: 1) image analysis to detect changes from bi-temporal remote sensing data; 2) simulation of flood and debris flow to synthesize training data of target variables (i.e., maximum water level and topographic deformation); 3) binarization of change information to link numerical simulation and remote sensing (or synthetic and real data); and 4) regression of target variables from a binary change map and DEM based on CNNs. Fig.~\ref{methodology} provides an overview of our methodology's concept.

The second and third modules deductively create output and input of training samples, respectively, for various artificial scenarios of floods and debris flows. The fourth module learns a nonlinear mapping inductively to solve the inverse problem from binary change information together with DEM to the maximum water level and topographic deformation. In a real scenario, the outcome of the first module is used as the input in the inference phase of the fourth module. The following subsections detail the four modules.

\subsection{Image Analysis}
There are various approaches for flood and landslide (including debris flow) detection that use either optical or SAR data, as previously reviewed in Section II. In this work, we select one of the simplest methods, based on spectral indices derived from bi-temporal optical images, to ease the third module and automate the whole processing chain. We use the normalized difference vegetation index (NDVI) if a near-infrared band is available. If only an RGB image is available, which is often the case for airborne emergency observation, we use the visible atmospherically resistant index (VARI). NDVI and VARI are calculated as follows
\begin{eqnarray}
    \textnormal{NDVI} & = & (\textnormal{NIR} - \textnormal{Red})\; / \;(\textnormal{NIR} + \textnormal{Red}) \\
    \textnormal{VARI} & = & (\textnormal{Green} - \textnormal{Red}\; / \;(\textnormal{Green} + \textnormal{Red} - \textnormal{Blue}) \nonumber
\end{eqnarray}

We calculate either NDVI or VARI from pre- and post-disaster optical images and detect areas where vegetation coverage decreases with changes due to floods and debris flows. Hard-thresholding is used in this work, and the threshold values to judge whether a pixel is vegetated or not are empirically set to 0.7 for NDVI and 0 for VARI. We assume that debris flows occur in vegetated mountainous areas. Note that the method used cannot detect the inundation of non-vegetation areas and also narrow debris flows occluded by tree crowns. Therefore, the change detection result provides only partial information on the flood and debris flow extent areas, with possible missing (i.e., false negatives). Regression models (Section III-D) learn to inpaint such missing information from synthetic data created by simulation (Section III-B) and binarization (Section III-C).

\begin{figure*}[t]
  \centering
  \includegraphics[width=\linewidth]{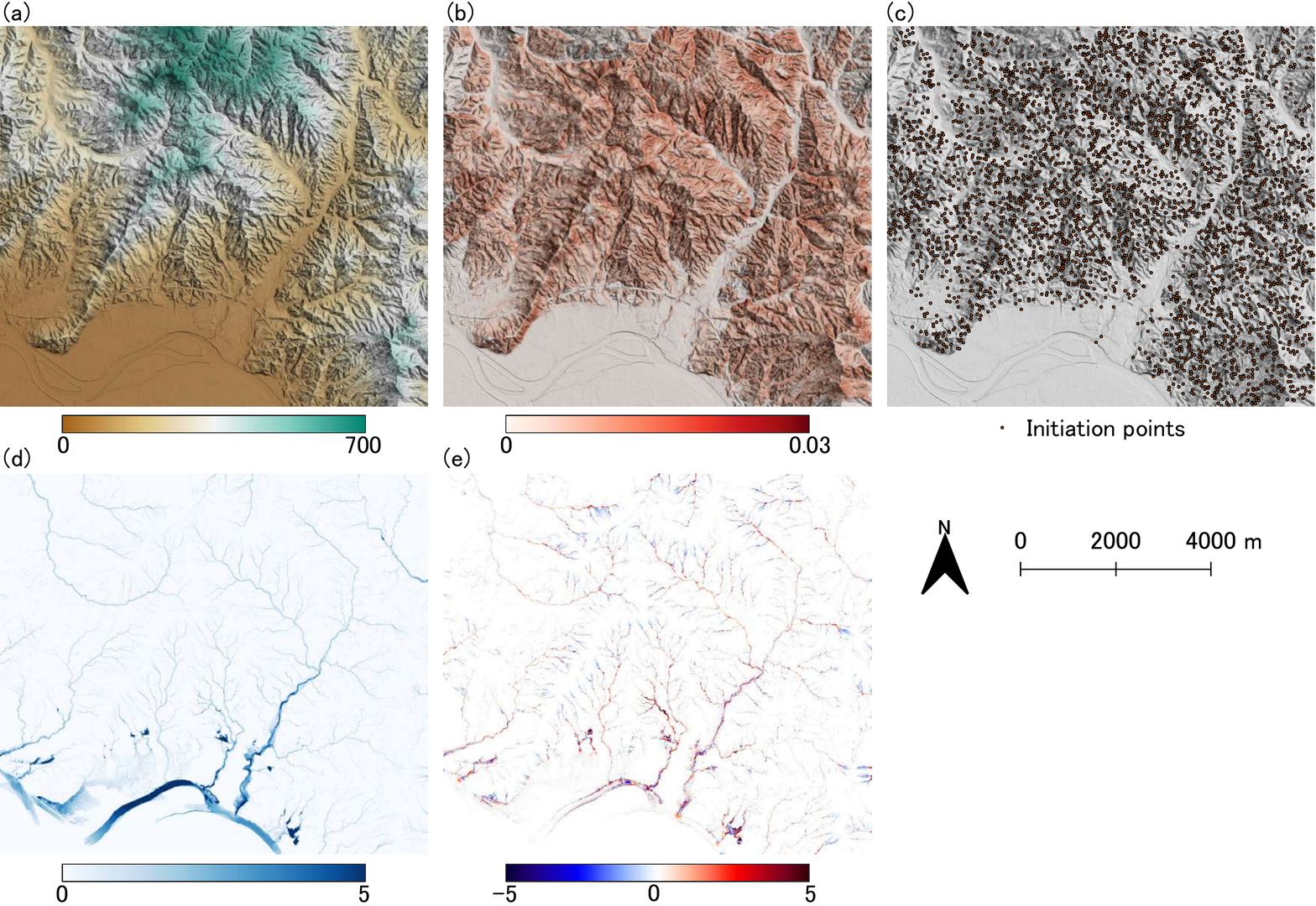}
  \caption[]{Input data for the simulation and one example of the output. (a) is elevation of the target region, (b) is distribution of the probability to be initiation points obtained by the logistic regression, (c) is an example of the initiation points generated by (b) and pseud numbers, (d) and (e) are an example of results on case 54 for maximum water level and terrain deformation, respectively. The size of points is enlarged to be displayed in (c) \cite{yamanoi2020preprint}.}
  \label{sim}
\end{figure*}

\begin{figure*}[t]
\begin{tabular}{cc}
\begin{minipage}{0.5\hsize}
\begin{center}
\includegraphics[scale=0.5]{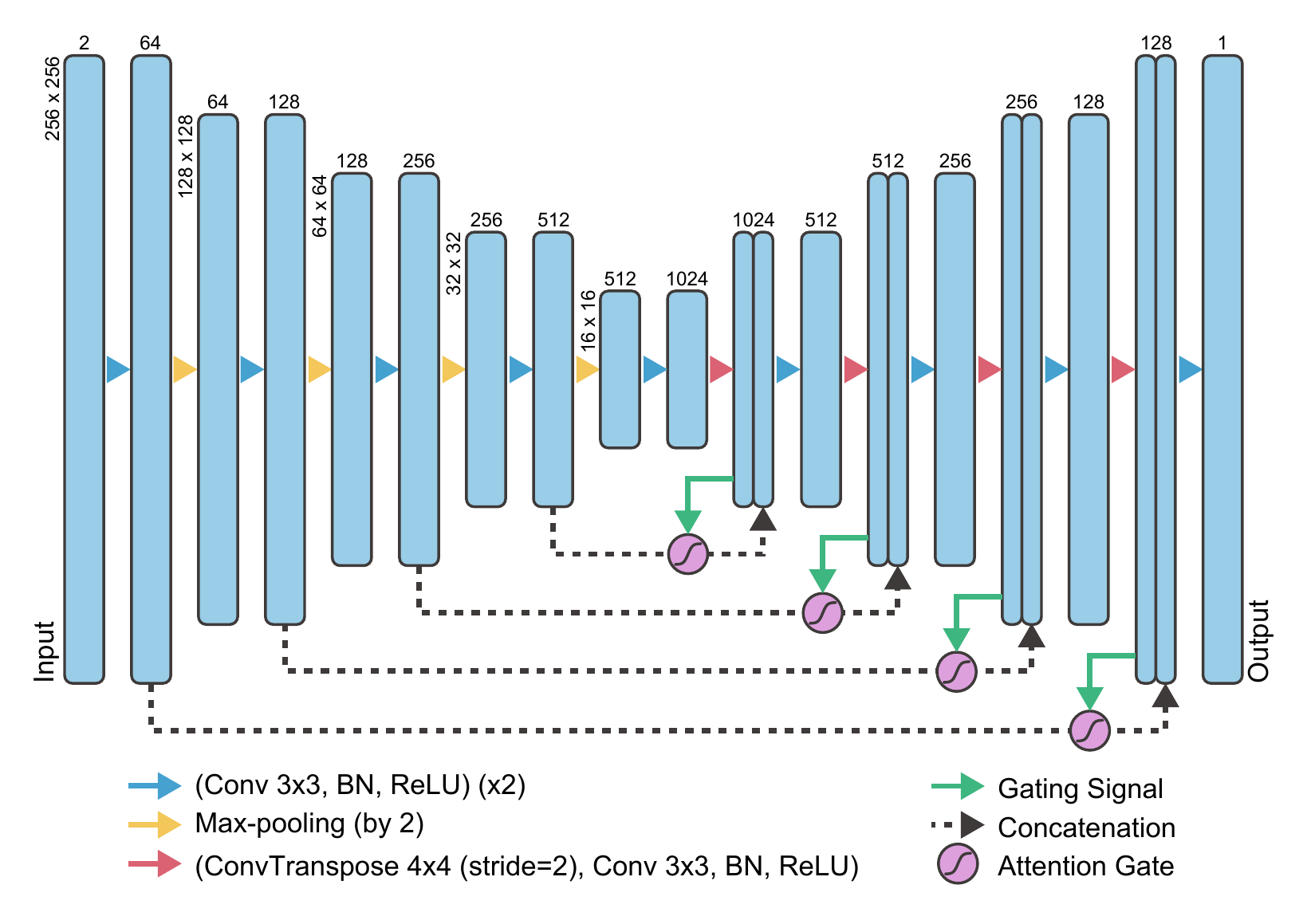}
\end{center}
\vspace{-5mm}
\subcaption{Attention U-Net}
\label{fig:no-1}
\end{minipage}
\begin{minipage}{0.5\hsize}
\begin{center}
\includegraphics[scale=0.5]{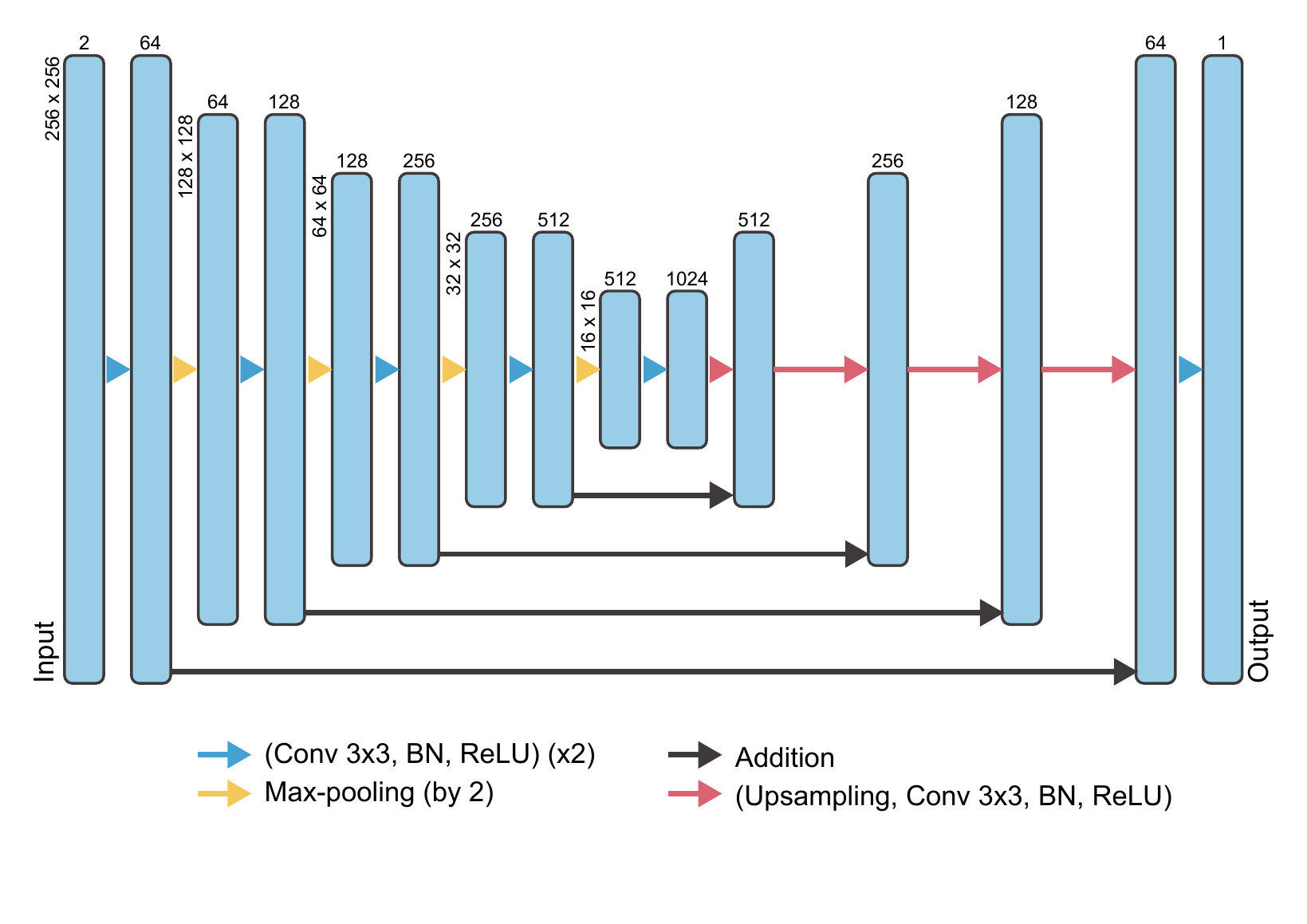}
\end{center}
\vspace{-5mm}
\subcaption{LinkNet}
\label{fig:no-2}
\end{minipage}
\end{tabular}
\caption{Architectures of investigated (a) Attention U-Net and (b) LinkNet.}
\label{architecture}
\end{figure*}

\subsection{Numerical Simulation}
In this work, the simulation methods developed by \cite{Takahashi2007} were used. Dynamics of the debris flow can be described by the governing equations, based on shallow water equations that take erosion and deposition processes into consideration. When erosion takes place, the water and sediment at the ground/river bed are retrieved into the flow body. On the other hand, both the sediment and water are trapped in the bed when deposition takes place. To express these processes, the following equations are employed.

\begin{equation}
      \frac{\partial {\bf U}}{\partial t}
      +\frac{\partial {\bf E}}{\partial x}
      +\frac{\partial {\bf F}}{\partial y}
      ={\bf S},
      \label{eq:deb1}
\end{equation}

\scriptsize
\begin{eqnarray}
  {\bf U} & = & \left(
  \begin{array}{c}
  h \\
  uh \\
  vh \\
  Ch \\
  z_b
  \end{array}
  \right),
  {\bf E} = \left(
  \begin{array}{c}
  uh \\
  u^2h+\frac{1}{2}gh^2 \\
  uvh \\
  Cuh \\
  0
  \end{array}
  \right),
  {\bf F} = \left(
  \begin{array}{c}
  vh \\
  uvh \\
  v^2h+\frac{1}{2}gh^2 \\
  Cvh \\
  0
  \end{array}
  \right), \nonumber\\
  % \label{eq:deb2}
  {\bf S} & = & \left(
  \begin{array}{c}
  i \\
   gh\left( S_{0x}-S_{fx}  \right) + \frac{\partial}{\partial x}\left\{ \epsilon \frac{\partial \left( uh \right)}{\partial x}\right\}  + \frac{\partial}{\partial y}\left\{ \epsilon \frac{\partial \left( uh \right)}{\partial y} \right\} \\
   gh\left( S_{0y}-S_{fy}  \right) + \frac{\partial}{\partial x}\left\{ \epsilon \frac{\partial \left( vh \right)}{\partial x}\right\}  + \frac{\partial}{\partial y}\left\{ \epsilon \frac{\partial \left( vh \right)}{\partial y} \right\} \\
  iC_* \\
  -i,
  \end{array}
  \right),
  % \label{eq:deb3}
  \label{eq:deb2}
\end{eqnarray}
\normalsize
where \({\bf U}\) , \({\bf E}\) , \({\bf F}\) , and \({\bf S}\)  are the conservative variable, flux for x- and y- directions, and source vectors, respectively.
The term \( h \) is the flow depth; \( u \) , \( v \)  are the velocity of x- and y- directions, respectively; \( C \)  is the sediment concentration of flow body; \( z_b \)  is the river/ground bed elevation; \( g \)  is the gravity acceleration; and \( \epsilon \)  is the eddy momentum diffusivity.
The terms \( S_{0x} \) and \( S_{0y} \) are the topographical gradients for x- and y-directions, respectively.
The terms \( S_{fx} \) and \( S_{fy} \) are the frictional gradients for x- and y- directions, respectively, which are calculated by different equations for three flow modes, stony debris flows, hyper-concentrated flow, and water flows, depending on the concentration of flow body \( C \) \cite{Takahashi1991}.
The term $i$ is the erosion/deposition velocity, which is calculated by the balance of the equilibrium concentration obtained by the function of water surface gradients.
It is also calculated by different functions for the three flow modes.
Additionally, in this study, fluidization rate  \( \gamma \) is introduced to consider the transformation of the fine solid material to fluid. The effective specific weight of fluid material \(\rho\) and its concentration in deposited material \( C_* \) are modified by the equation below.
\begin{equation}
    \rho = \frac{\gamma \sigma C_{*0} + \rho_0 \left( 1 - C_{*0} \right)}
    {\gamma C_{*0} + \left( 1 - C_{*0}\right)},
    C_*=C_{*0}\left( 1-\gamma \right),
    \label{eq:spc}
\end{equation}
where \(C_{*0}\) is the original sediment concentration in the deposited material and \(\rho_0\) is the specific weight of water.

For numerical modeling, we used the MacCormack scheme with artificial viscosity, which is categorized into a two-step scheme in FDM (finite-difference methods).
The code is parallelized by employing both MPI and OpenMP and therefore can be simulated on large-scale supercomputers \cite{yamanoi2020preprint}.

The location of the initiation points of a debris flow (e.g., Fig.~\ref{sim}(c)) is required in order to conduct the simulation, and can be obtained only after the disaster event.
The statistically-predicted initiation points can substitute for the actual data; however, the predicted damage is not uniquely derived \cite{yamanoi2020preprint}.
Therefore, the method can generate many possible damage results.

In this method, in order to generate the artificial damage data, we use the distribution of a probability, which is obtained by logistic regression employing the actual disaster data and topographical data.
In the regression, local slopes, accumulation, and tangential- and plan-curvatures are selected as explanatory variables.
The obtained probability is shown in Fig.~\ref{sim} (b).
In order to use this data for simulation inputs, sets of pseudo-random numbers were used to convert to the binary point distribution shown in Fig.~\ref{sim} (c) as an example.
From elevation (Fig.~\ref{sim} (a)) and points (Fig.~\ref{sim} (c)), the simulation calculates the transport of debris and water flow temporally and spatially.
We select the maximum water level and final terrain deformation as the target variables, which represent the damage from the hazards, as shown in Fig.~\ref{sim} (d) and (e).
%By this presented method, \textcolor{red}{for Northern Kyushu 2017, } we generated 60 sets of the input points by changing the random seed, and conducted simulations for the 60 cases simultaneously using K computer installed in RIKEN Center for Computational Science, Japan.
%\textcolor{red}{For Western Japan 2018, we generated ten sets of the input points as well, and conducted simulation changing the ten values of \(\gamma\) from 0 to 0.6 for each set of input points, therefore 100 results were generated.}

\subsection{Binarization of Change Information}
One key objective of the proposed methodology is to ensure transferability of regression models trained on synthetic data to real data. Based on physical reasoning behind the change detection method presented in Section III-A, we attempt to create synthetic binary change maps that resemble a real change detection map obtained by remote sensing image analysis. In addition, inspired by domain randomization~\cite{tobin2017domain}, we try to make the distribution of the synthetic binary data sufficiently wide and varied so that regression models trained on the synthetic data work robustly with real data.

The binary change map obtained from the observation contains false positives and false negatives due to the characteristics of the detection method and the observed data, and so it is necessary to perform binarization of change information (i.e., maximum water level and topographic deformation) simulating these errors. When using the change detection method based on vegetation-related spectral indices from optical images, it should be noted that floods and debris flows are not detected in areas that satisfy one of the following two conditions: 1) areas with low NDVI before the disaster, and 2) areas where occlusion occurs due to the effects of tree crowns and incident angles. In the binarization, the former can be easily synthesized by using the NDVI/VARI image derived from the pre-disaster optical imagery. In the latter case, more detailed three-dimensional information is required to perform model-based simulation. For simplicity, we perform morphological erosion to synthesize false negative pixels. Pixels having decreases in vegetation due to other reasons than a disaster are erroneously recognized as flood or debris flow areas. Since it is difficult to synthesize such phenomenon based on a model, we randomly add noise to synthesize false positive pixels. Furthermore, the simulation results include floods and debris flows everywhere, and there are very few negative examples in which no flood or debris flow has occurred. We use a cut out~\cite{devlin2018bert} for data augmentation to enforce regression models to output 0 if there is no change.

\begin{figure*}[t!]
\begin{tabular}{cc}
\begin{minipage}{0.5\hsize}
\begin{center}
\includegraphics[width=9.0cm]{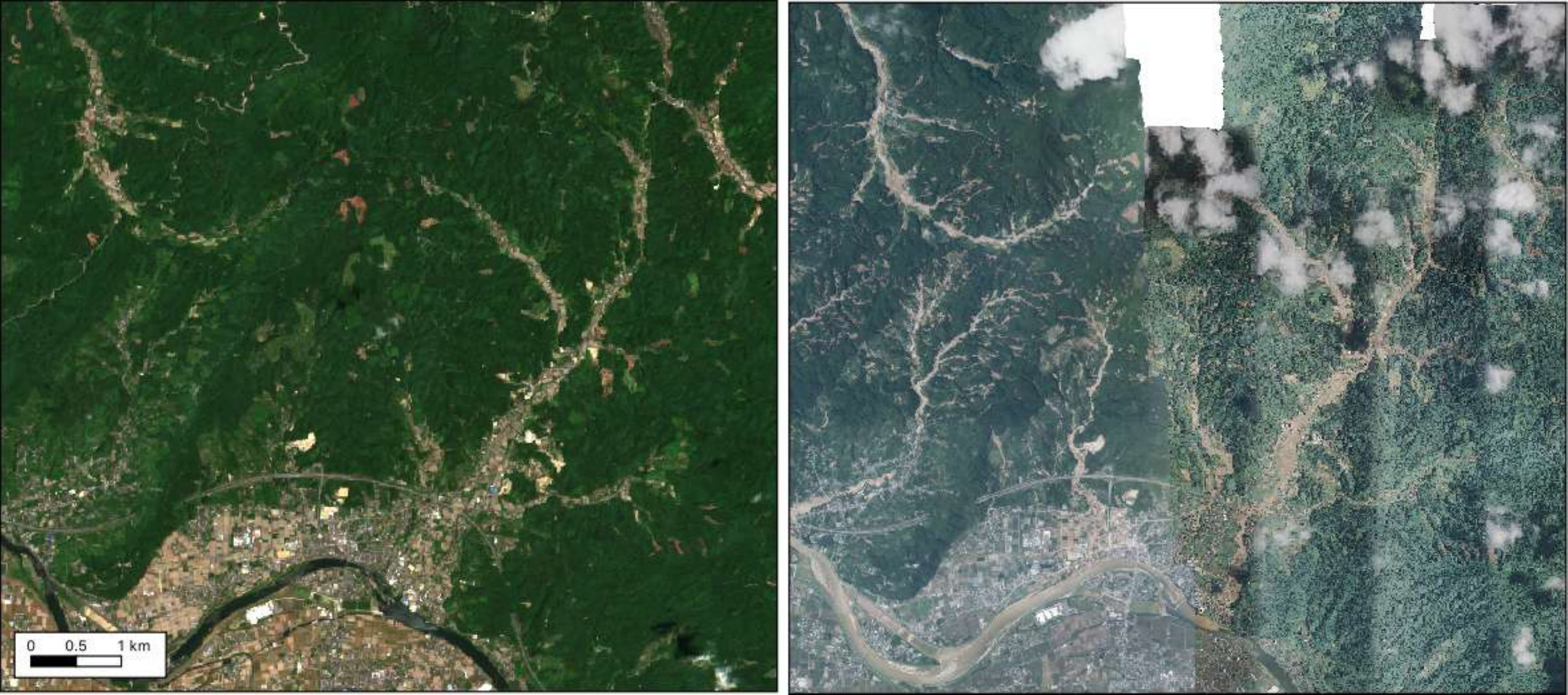}
\end{center}
\vspace{-2mm}
\subcaption{Northern Kyushu 2017}
\label{studyarea1}
\end{minipage}
\begin{minipage}{0.5\hsize}
\begin{center}
\includegraphics[width=9.0cm]{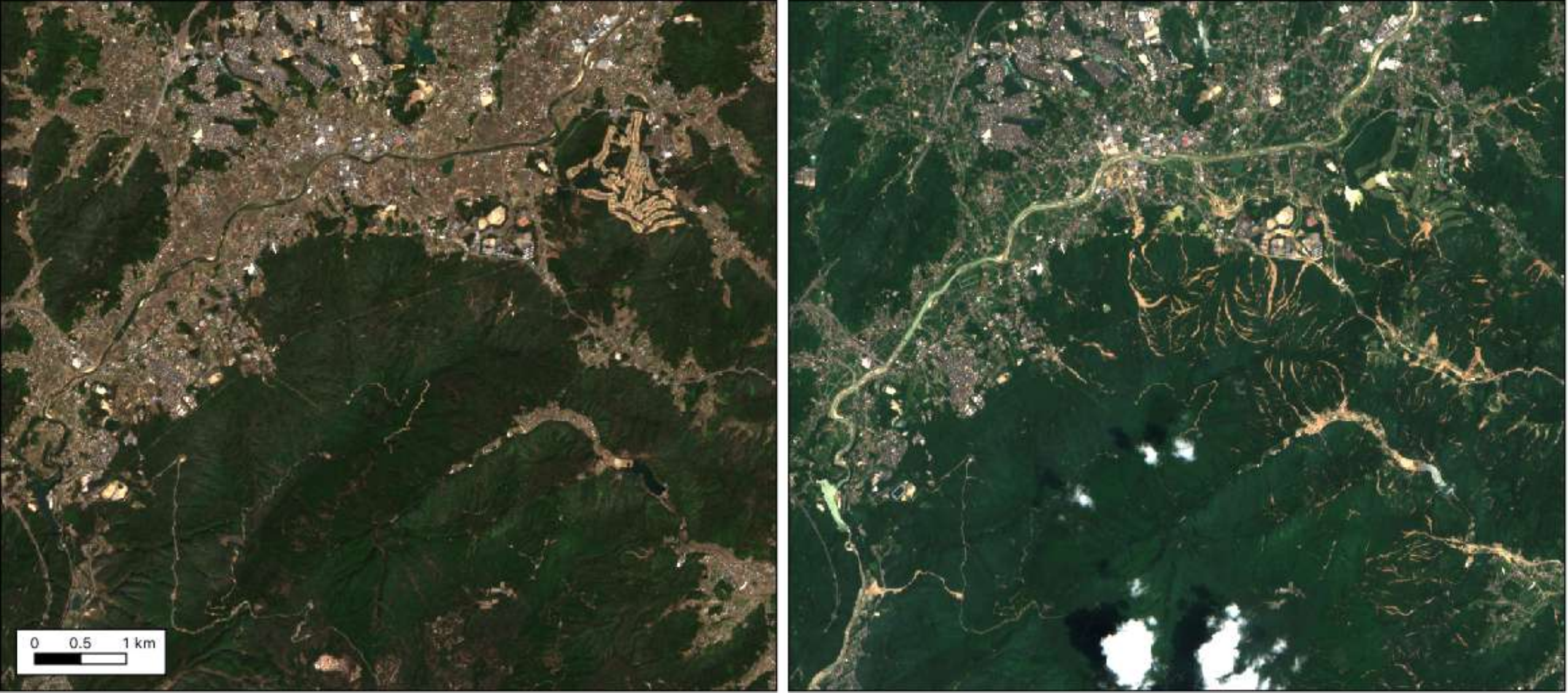}
\end{center}
\vspace{-2mm}
\subcaption{Western Japan 2018}
\label{studyarea2}
\end{minipage}
\end{tabular}
\caption{(a) Pre-disaster Sentinel-2 imagery (left) and post-disaster aerial RGB imagery (right) over the study area of Northern Kyushu 2017. (b) Pre- (left) and post-disaster (right) Sentinel-2 images over the study area of Western Japan 2018.}
\label{studyarea}
\end{figure*}

\subsection{CNN-based Regression}
By estimating the maximum water level and tomographic changes separately by individual CNNs, we are able to account for their different physical properties.

Regression models learn a nonlinear mapping $f_\theta$ from input $x$, which is composed of binary change map and slope images, to output $y$ (maximum water level or topographic deformation): $f_\theta: x \rightarrow y$.

The smooth $L_1$ loss (or Huber loss) is used for robust regression of target variables:
\begin{equation}
\mathcal{L}_{\delta}(y,f_\theta(x))=
\begin{cases}
\frac{1}{2}(y-f_\theta(x))^2, & |y-f_\theta(x)| \le \delta \\
\delta|y - f_\theta(x)|-\frac{1}{2}\delta^2, & \mathrm{otherwise}.
\end{cases}
\end{equation}
The smooth $L_1$ loss combines the advantage of the $L_2$ loss (gradient decreases when the loss gets close to local minima) and the $L_1$ loss (less sensitive to outliers). We set $\delta$ to 1.

For the regression models, we investigate variations of two architectures, Attention U-Net~\cite{oktay2018attention} and LinkNet~\cite{chaurasia2017linknet}, which have consistently shown high performance in semantic segmentation~\cite{zhou2018d}. We adopt model fusion to further improve accuracy by taking the average of the outputs of the two models as the final output. Fig.~\ref{architecture} illustrates the architectures of Attention U-Net and LinkNet investigated in this work with the size of feature maps. The size of input images is $256 \times 256$ pixels.

U-Net is an encoder-decoder structure with skip connections, which shares the information learned by the encoder with the decoder through concatenation~\cite{ronneberger2015u}. We adopt a modified version of the architecture proposed in~\cite{oktay2018attention}, which incorporates a self-attention mechanism in U-Net with contextual information extracted at a coarser scale. Each encoder block is composed of two convolutionsal layers, each followed by batch normalization and a rectified linear unit (ReLU).

The original LinkNet also has an encoder-decoder structure with residual blocks and skip connections, but it shares the information learned by the encoder with the decoder through additive operations. In this work, we do not use residual blocks, but use the same one as Attention U-Net, which leads to a more compact model.

We use the Adam solver for optimization with a learning rate of 0.0001. Xavier initialization is used to initialize the weights. The batch size is 32 and the number of epochs is 200.

\begin{table*}[t]
\renewcommand{\arraystretch}{1.2}
\caption{Numerical evaluation results in synthetic data experiments.}
\label{tab:synthetic}
\centering
\begin{tabular}{c|c|c|c|c|c|c|c|c}
\hline
 & \multicolumn{4}{c|}{Northern Kyushu 2017} & \multicolumn{4}{c}{Western Japan 2018} \\
\cline{2-9}
Approach & \multicolumn{2}{c|}{Water Level} & \multicolumn{2}{c|}{Topographic Def.} & \multicolumn{2}{c|}{Water Level} & \multicolumn{2}{c}{Topographic Def.} \\
\cline{2-9}
             & RMSE        & LSHI        & RMSE        & LSHI        & RMSE        & LSHI        & RMSE        & LSHI        \\
\hline
Average Sim. & 0.1567      & 0.9602      & 0.1499      & 0.8746      & 0.5214      & \bf{0.9772} & 0.3585      & \bf{0.9701} \\
Att. U-Net   & 0.1322      & \bf{0.9674} & 0.1302      & \bf{0.9222} & 0.2837      & 0.9727      & 0.3193      & 0.9132      \\
LinkNet      & 0.1293      & 0.9552      & 0.1319      & 0.9183	     & 0.3007      & 0.9312      & 0.3425      & 0.8992      \\
Fusion       & \bf{0.1246} & 0.9652      & \bf{0.1246} & 0.9191      & \bf{0.2784} & 0.9546      & \bf{0.3161} & 0.9176      \\
\hline
\end{tabular}
\end{table*}

\section{Experiments}
In this section, we demonstrate the usefulness of the proposed framework for estimating the maximum water level and topographic deformation, using data from two complex disasters where floods and debris flows occurred simultaneously. The performance of the proposed CNN-based methods---Attention U-Net, LinkNet, and the fusion---is evaluated quantitatively and qualitatively with both synthetic and real data experiments.

\subsection{Datasets}
The experiments focus on the analysis of floods and debris flows caused by two disaster events: 1) torrential rain in July 2017 in northern Kyushu, Japan; and 2) torrential rain in July 2018 in western Japan. Hereinafter, we refer to datasets for these events as \textit{Northern Kyushu 2017} and \textit{Western Japan 2018}, respectively. Fig.~\ref{studyarea} shows the study scenes for the two events. Each of them covers \SI{8}{\kilo\meter}  $\times$ \SI{9}{\kilo\meter} with $1600 \times 1800$ pixels at a ground sampling distance of \SI{5}{\meter}, including a variety of geographically different areas (e.g., urban, agriculture fields, mountain forests). We used the DEM released by GSI for the simulation. Details of each dataset are given below.

\subsubsection{Northern Kyushu 2017}
Torrential rains hit northern Kyushu on July 5 and 6, 2017 and damaged many houses (336 were completely destroyed, 1096 partially destroyed) and caused human casualities. In this work, we analyze the area surrounding the city Asakura, where debris flows and inundation occurred simultaneously. %The study scene covering $8 km \times 9 km$ includes a variety of geographically different areas (e.g., urban, agriculture fields, mountain forests) as shown in Fig.~\ref{studyarea}.

For remote sensing images, we use a pre-disaster Sentinel-2 image and aerial photographs released by the Geospatial Information Authority of Japan (GSI) soon after the disaster (Fig.~\ref{studyarea}(a)). The mosaic aerial imagery was made from two different observations. We use a debris-flow and inundation extent map manually created by experts from aerial photographs as reference data for evaluation. Clouds and missing areas included in the aerial photographs used in the analysis were masked out manually and were not included in the evaluation.

By the simulation method presented in Section II-B, we generated 60 sets of input points by changing the random seed, and conducted simulations for the 60 cases simultaneously using the K computer installed in the RIKEN Center for Computational Science in Japan.
%We generated a synthetic dataset composed of 60 simulation cases as described in Section II-B.
By using the reference map of the flood and debris flow extent map created by the visual interpretation as the input of the simulation, we obtained the maximum water level and topographic deformation with high accuracy and used it as reference data for evaluation. We refer to this simulation result as the \textit{reference simulation}.

\subsubsection{Western Japan 2018}
The second dataset was collected before and after the floods and debris flows caused by torrential rains in western Japan during the period June 28 to July 8, 2018. These rains caused river flooding, inundation, flash floods, and debris flows in many areas of western Japan, with the death toll exceeding 200. In this work, we analyze an area in Higashihiroshima, which was severely affected by the floods and debris flows.

We use pre- and post-disaster Sentinel-2 images (Fig.~\ref{studyarea}(b)) as input for remote sensing image analysis. A debris-flow extent map created by experts Evisual interpretation from aerial photographs is used as reference for evaluation of disaster extent detection. For the Western Japan 2018 dataset, pre- and post-disaster LiDAR derived digital terrain models (DTMs) are also available for the study area. The difference in the DTMs is used as reference for numerical evaluation of topographic deformation. Note that we mask out (or set to zero) the values of the topographic deformation reference at unchanged pixels in the debris-flow extent reference map.

We generated ten sets of input points and conducted the simulation, changing the ten values of \(\gamma\) from 0 to 0.6 (i.e., \(\gamma = 0, 0.01, 0.02, 0.05, 0.1, 0.2, 0.3, 0.4, 0.5, 0.6 \)) for each set of input points; therefore 100 results were generated.
%A total of 100 simulation cases were generated by performing 10 randomly generated sets of debris flow start points for each of 10 different parameter sets \textcolor{red}{(i.e. \(\gamma = 0, 0.01, 0.02, 0.05, 0.1, 0.2, 0.3, 0.4, 0.5, 0.6 \))}.
By using the debris-flow initiation points manually annotated by experts from Hiroshima University for the input of the simulation, we obtained 10 reference simulations of the maximum water level and topographic deformation for the 10 parameter sets. We quantitatively evaluate the accuracy of the 10 reference simulations of topographic deformation in comparison with the LiDAR-derived reference and use the best one as the reference simulation. The accuracy of the reference simulation for topographic deformation is discussed in Section IV-D.

%\begin{figure}[t]
%\centering
%\includegraphics[width=\linewidth]{./Fig/StudyImages.png}
%\caption{(Left) Pre-disaster Sentinel-2 imagery and (right) %post-disaster aerial RGB imagery over the study area.}
%\label{studyarea}
%\end{figure}

\subsection{Evaluation Metrics}
We use three metrics for quantitative evaluation: 1) the root-mean-square error (RMSE), 2) the intersection over union (IoU), and 3) the log-scaled histogram intersection (LSHI).

The pixel-wise accuracy of the predictions obtained by the proposed method is evaluated by RMSE using the reference simulation of maximum water level and topographic deformation. We use RMSE in both synthetic and real data experiments. In the real data experiment with the Western Japan 2018 dataset, RMSE is computed for topographic deformation using the LiDAR-derived reference, which is denoted as \textit{RMSE (real)}.

For evaluating the accuracy of detecting affected areas, we calculate IoU by comparing the binary change map obtained by simple thresholding from the maximum water level and topographic deformation with the reference map of the flood and debris flow extent map obtained by visual interpretation. IoU is used only in the experiments with real data where visual interpretation results are available. We adopt the best threshold for all results for fairness.

In addition to the spatial (pixel-by-pixel) details of the disaster, it is also important to understand the overall scale of the disaster. We adopt LSHI to measure the accuracy in terms of the overall scale of flood and debris flow. We calculate LSHI in both synthetic and real data experiments. As with RMSE, LSHI is computed for topographic deformation using the LiDAR-derived reference for the Western Japan 2018 data, which is denoted as \textit{LSHI (real)}.

For the comparison of RMSE and LSHI, the baseline is the average of all simulation outcomes for each set of parameters, which can be regarded as the expected value of target variables with Monte Carlo simulations. For the Western Japan 2018 dataset, we have the 10 average simulations and use the best one as the baseline for each test case. Note that the average simulation is more accurate than any individual simulation. For IoU, the binary change detection result of remote sensing image analysis and the reference simulation are also compared. For RMSE (real) and LSHI (real) of the topographic deformation in the Western Japan 2018 real data experiment, we evaluate the reference simulation as well.

\begin{figure*}[t!]
\begin{tabular}{cc}
\begin{minipage}{0.5\hsize}
\begin{center}
\includegraphics[width=9.0cm]{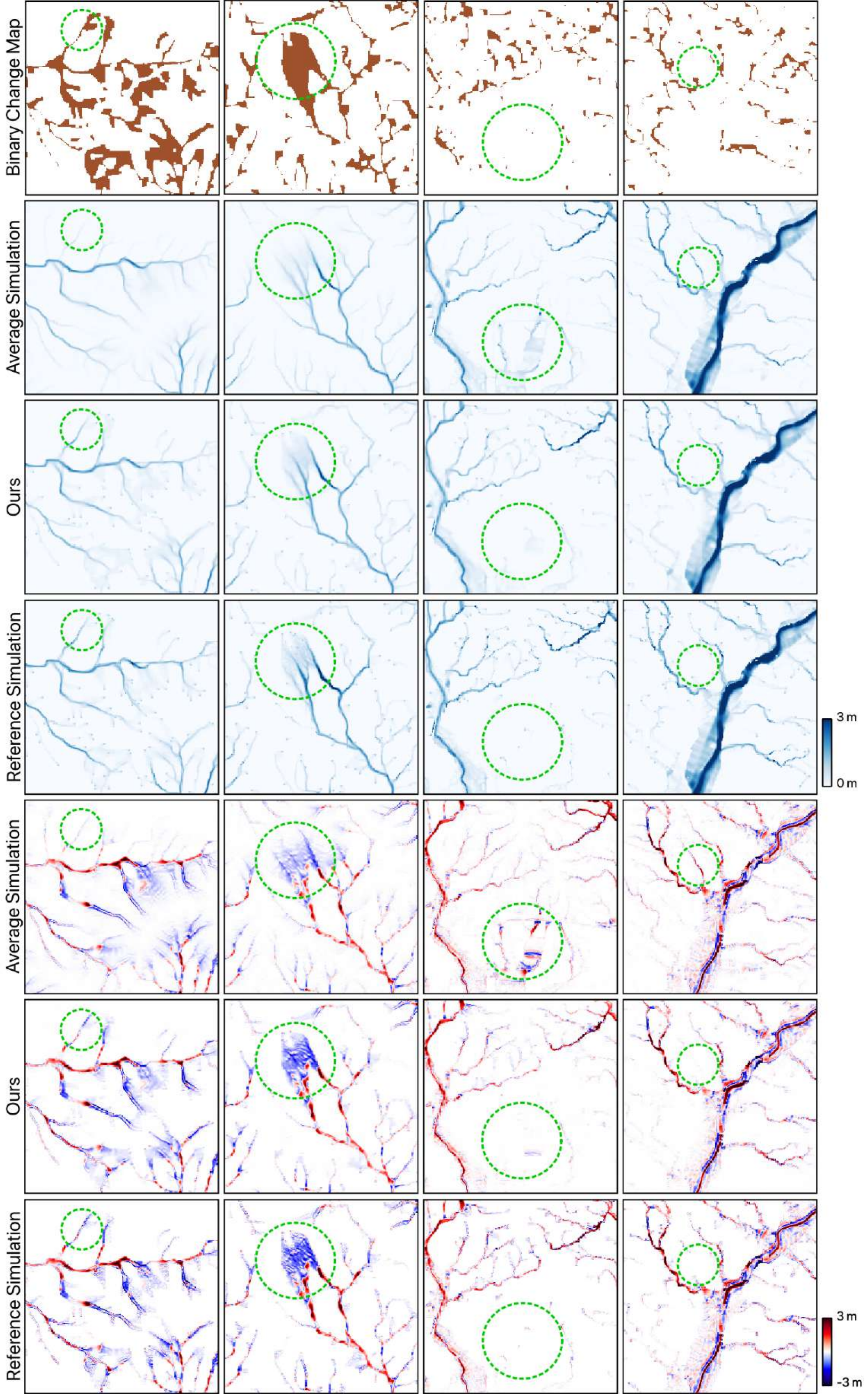}
\end{center}
\vspace{-2mm}
\subcaption{Northern Kyushu 2017}
\label{fig:synthetic:a}
\end{minipage}
\begin{minipage}{0.5\hsize}
\begin{center}
\includegraphics[width=9.0cm]{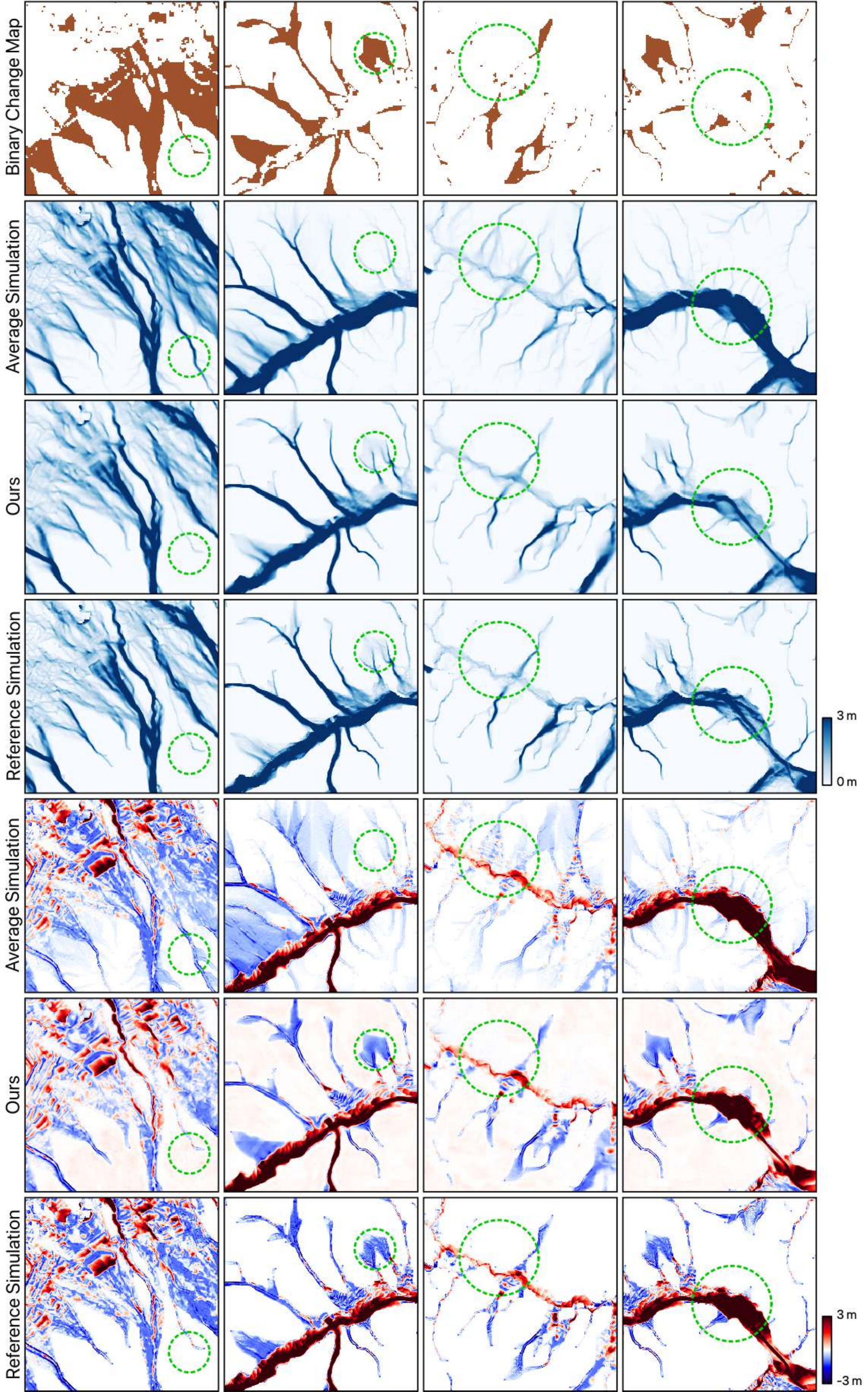}
\end{center}
\vspace{-2mm}
\subcaption{Western Japan 2018}
\label{fig:synthetic:b}
\end{minipage}
\end{tabular}
\caption{Examples of the binary change map (first row), the average simulation, our results, and the reference simulation for both maximum water level (second to fourth rows) and topographic deformation (fifth to seventh rows) for (a) Northern Kyushu 2017 and (b) Wstern Japan 2018.}
\label{fig:synthetic}
\end{figure*}

\subsection{Synthetic Data Experiment}
In the synthetic data experiments, we evaluate the generalization ability of the regression models for unseen data.
We split the simulation cases into training and test sets as follows.
%the following two ways.
\begin{itemize}
    \item \textbf{Northern Kyushu 2017}: We use 40 cases for training and 20 cases for testing; 1750 patches with non-overlapping sampling are used for training.
    \item \textbf{Western Japan 2018}: We use 70 cases for training and 30 cases for testing; 3010 patches with non-overlapping sampling are used for training.
    %\item Scenario-wise split: Training and test data are separated with respect to different simulation scenarios. We use 40 scenarios for training and 20 scenarios for testing. 1750 patches with non-overlapping sampling are used for training.
    %\item Region-wise split: We use 70\% of the western part of the entire scene as training data and 30\% of the eastern part as test data. 1813 patches of non-overlapping sampling are used for training.
\end{itemize}
%In the scenario-wise split,
The average simulation of the training data is compared as the baseline. %We make comparisons between only different models in the region-wise split, since the average simulation is not available for the test region.

\subsubsection{Quantitative Results}
Table~\ref{tab:synthetic} shows the accuracy of the estimated maximum water level and topographic deformation calculated by the regression models based on Attention U-Net, LinkNet, and their fusion.
%In the scenario-wise split,
All the models outperform the average simulation in RMSE for both datasets, indicating that the networks successfully learn a nonlinear mapping with respect to different binary change maps rather than outputting a simple average of training data. The superiority of the proposed methods in RMSE are more evident with the Western Japan 2018 dataset.
One possible reason for the large RMSE of the average simulation in the Western Japan 2018 dataset is that the average simulation, which is the sample average, deviates from the true expected value since only about seven cases are included in the training data for each parameter set.
The RMSE values of our results for the Western Japan 2018 dataset are larger than those for the Northern Kyushu 2017 dataset due to the large variation in the simulation results.
%This is because in the Western Japan 2018 dataset, the simulation results based on the different scenarios largely vary and each simulation result is far from the average of the training data. The large variation in the simulation results for the Western Japan 2018 dataset can be also indicated by the fact that the errors of our results are larger than those for the Northern Kyushu 2017 dataset.
The average simulation shows better LSHI values for Western Japan 2018. The reason for this result is that the average simulation (the best among the 10 average simulations) uses the same parameters as the test case and the scale of the disaster is very similar to each test case, while the regression model uses all the training data generated with the different parameters and the scale of the disaster is not optimized.
%The estimation accuracy is worse for the region-wise split than that for the scenario-wise split. The results suggest that the learned models are less transferable between different spatial/geographical regions than between different simulation scenarios. This is reasonable because the binarization step differs based on pre-disaster land covers, and thus the region-wise split results in larger domain gap between training and test sets.

Among the different CNN-based regression models, the fusion method shows the best performance in RMSE for both datasets. For LSHI, Attention U-Net outperforms LinkNet for both target variables with the two datasets, and thus the model fusion is not helpful for improving LSHI.

\subsubsection{Qualitative Results}
Fig.~\ref{fig:synthetic} shows four example patches of the binary change map of the input, the average simulation, our result with model fusion, and the reference simulation used as ground truth for both maximum water level and topographic deformation. Our results resemble the reference simulation and outperform the average simulation by exploiting the hints of disaster locations from the binary change map.
The effectiveness of the proposed fusion method compared to the average simulation is noticeable in the places indicated by the green circles in the figure. False positives and false negatives are inevitable for the average simulation since it only represents the expected value of the target variable based on the Monte Carlo simulation, whereas the proposed fusion method accurately estimates the target variables for each different scenario of the test set, based on the disaster locations included in the binary change detection map. The third and fourth columns in Fig.~\ref{fig:synthetic}(a) and all examples in Fig.~\ref{fig:synthetic}(b) clearly demonstrate that the proposed fusion method can estimate the target variables in the missing parts of the binary change map.

\begin{table*}[t]
\renewcommand{\arraystretch}{1.2}
\caption{Numerical evaluation results in real data experiments.}
\label{tab:real}
\centering
\resizebox{\textwidth}{!}{%
\begin{tabular}{c|c|c|c|c|c|c|c|c|c|c|c|c|c|c}
\hline
 & \multicolumn{6}{c|}{Northern Kyushu 2017} & \multicolumn{8}{c}{Western Japan 2018} \\
\cline{2-15}
Approach & \multicolumn{3}{c|}{Water Level} & \multicolumn{3}{c|}{Topographic Def.} & \multicolumn{3}{c|}{Water Level} & \multicolumn{5}{c}{Topographic Def.} \\
\cline{2-15}
               & RMSE       & IoU        & LSHI       & RMSE       & IoU        & LSHI       & RMSE       & IoU        & LSHI        & \makecell{RMSE \\(real)} & RMSE        & IoU         & \makecell{LSHI \\(real)}  & LSHI \\
\hline
Detection      & ---        & 0.3054     & ---        & ---        & 0.3054     & ---        & ---        & 0.3177     & ---         & ---         & ---         & \bf{0.3177} & ---         & ---      \\
Average Sim.   & 0.5649     & 0.2872     & 0.5119     & 0.2362     & 0.2378     & 0.7391     & \bf{0.6419}& 0.1024     & \bf{0.9256} & 0.5328      & \bf{0.3946} & 0.0889      & 0.6261      & \bf{0.8684} \\
Att. U-Net     & 0.3889     & 0.4433     & 0.7811     & 0.1697     & 0.3765     & \bf{0.8910}& 0.7186     & 0.3160     & 0.6586      & 0.2303      & 0.4006      & 0.3018      & \bf{0.7129} & 0.5137  \\
LinkNet        & 0.3899     & 0.4420     & \bf{0.7924}& 0.1688     & 0.3536     & 0.8404     & 0.7121     & 0.3141     & 0.7155      & 0.2276      & 0.4020      & 0.3003      & 0.6984      & 0.5070 \\
Fusion         & \bf{0.3852}& \bf{0.4454}& 0.7840     & \bf{0.1654}& \bf{0.3790}& 0.8575     & 0.7131     & \bf{0.3199}& 0.6854      & \bf{0.2266} & 0.4000      & 0.3045      & 0.6903      & 0.4987  \\
\hdashline
Reference Sim. & 0          & 0.4070     & 0          & 0          & 0.4073     & 0          & 0          & 0.2266     & 1           & 0.4497      & 0           & 0.2146      & 0.6576      & 1 \\
\hline
\end{tabular}%
}
\end{table*}

\subsection{Real Data Experiment}
In the real data experiments, for our proposed methods, we used all the synthetic data for training and the remote sensing-derived change detection maps as input for the inference. We again adopt the average simulation of the training data as the baseline of simulation results. Furthermore, the remote sensing-derived change detection maps are regarded as the baseline of the remote sensing results.

\subsubsection{Quantitative Results}
Table~\ref{tab:real} summarizes the quantitative evaluation results for the real data experiment. The proposed methods show the best accuracy for all the evaluation metrics with the Northern Kyushu 2017 dataset and also for IoU, RMSE (real), and LSHI (real) that are based on the real reference data for the Western Japan 2018 dataset. The results demonstrated that by integrating remote sensing, simulation, and deep learning, it is possible to extract physically semantic disaster information that cannot be obtained by either remote sensing image analysis or simulation alone.

In the Northern Kyushu 2017 experiment, the IoU scores achieved by our approach clearly outperform the change detection result of remote sensing image analysis, indicating that the synergistic use of simulation and deep learning successfully improved the detection of disaster extent areas. The fact that our IoU scores are comparable to or even better than those of the reference simulation suggests that high accuracy has been achieved for flood and debris-flow detection. The RMSE and LSHI of our results are better than those of the average simulation with large margins, which supports the effectiveness of the proposed framework.

In the Western Japan 2018 experiment, our IoU scores are comparable to that of the change detection by remote sensing image analysis because the improvement in (or inpainting of) false negatives was offset by the increase in false positives. RMSE and LSHI results in the Western Japan 2018 experiment must be carefully interpreted due to the inaccuracy of the reference simulation, as is apparent in its limited scores in RMSE (real) and LSHI (real) for topographic deformation. In other words, RMSE (real) and LSHI (real) are the most important metrics, and our method shows its superiority over the best-effort simulation.

\begin{figure*}[t]
\centering
\includegraphics[width=0.9\linewidth]{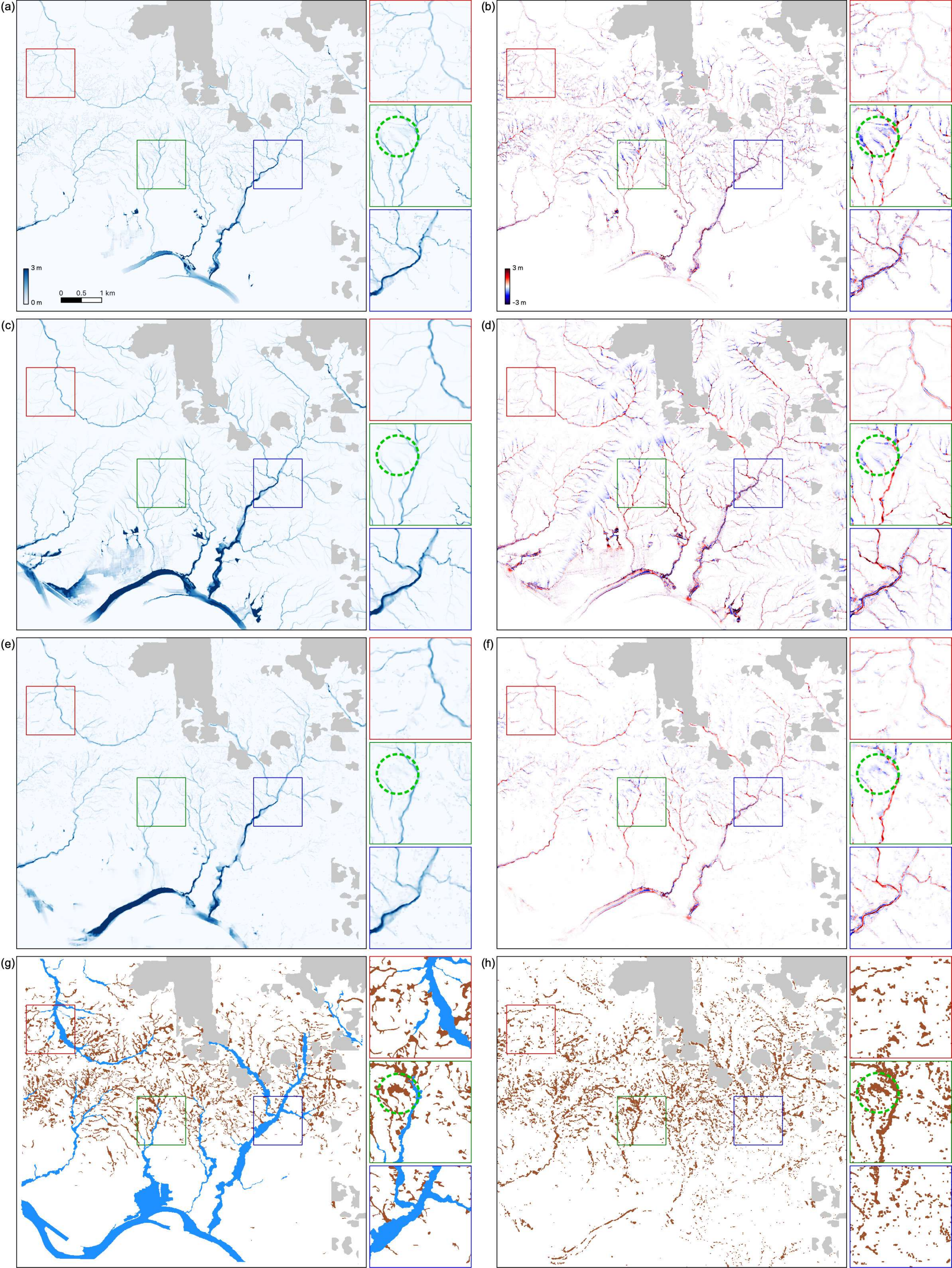}
\caption{(a)(c)(e) Maximum water level and (b)(d)(f) topographic deformation of (a)(b) reference simulation, (c)(d) average simulation, and (e)(f) our method with (g) reference map of (blue) flood and (brown) debris flow extent and (h) binary change detection map. Mask for clouds and no data is shown in gray.}
\label{realdataresults}
\end{figure*}

\begin{figure*}[t]
\centering
\includegraphics[width=0.9\linewidth]{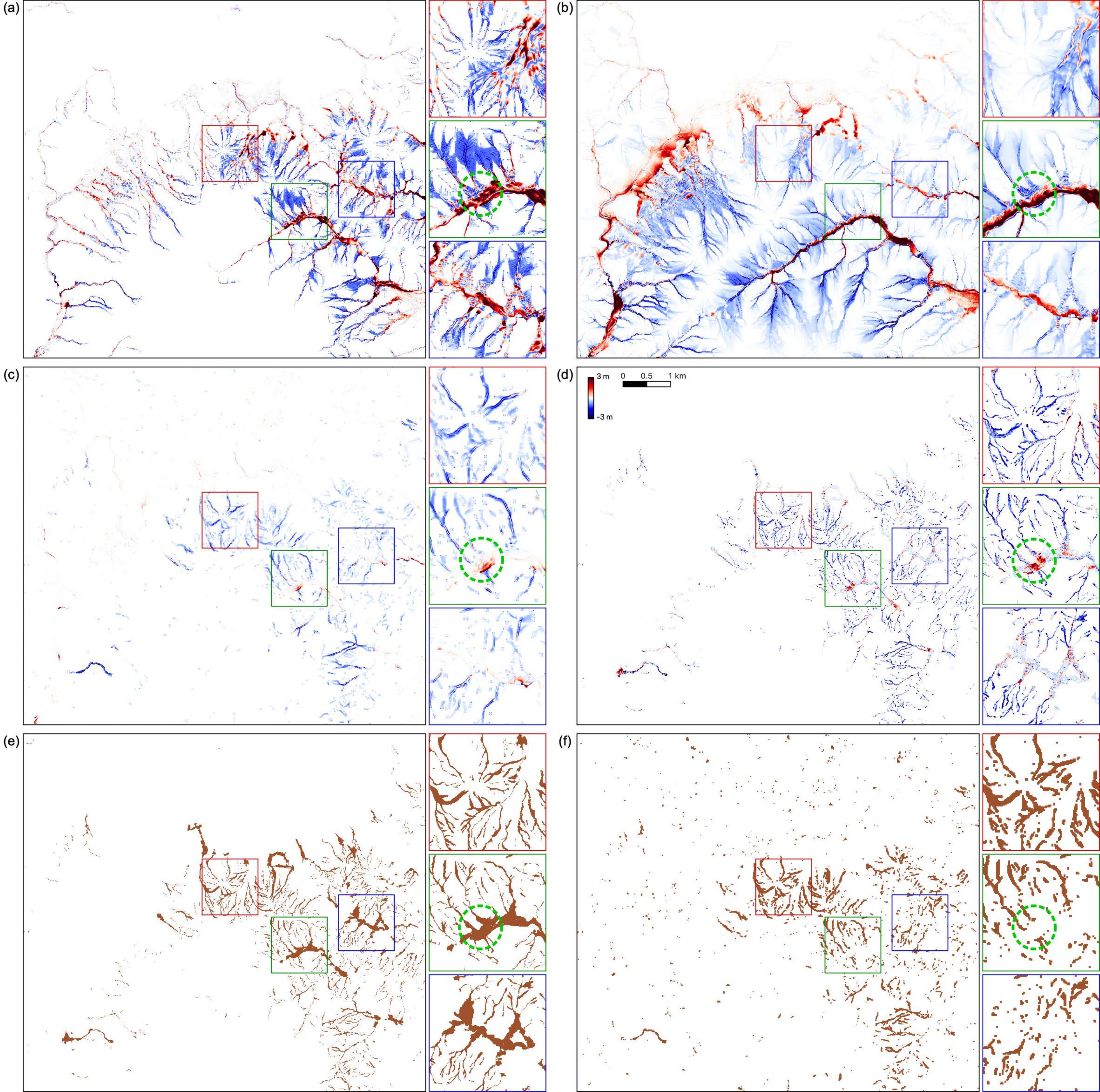}
\caption{Topographic deformation of (a) reference simulation, (b) average simulation, (c) our method, and (d) LiDAR-derived DTM with (e) reference map of (brown) debris flow extent and (f) binary change detection map.}
\label{hiroshima}
\end{figure*}

\subsubsection{Qualitative Results}
Fig.~\ref{realdataresults} shows the visual results of the Northern Kyushu 2017 experiment by comparing the reference simulation, the average simulation, and our estimation results of the maximum water level and topographic deformation together with the visual interpretation reference and the flood and debris-flow detection map obtained by image analysis. It can be visually observed that the results of the proposed fusion method achieve more accurate estimation than the average simulation, which is consistent with the quantitative evaluation results. The average simulation includes overestimation in the northern and southern regions, whereas the proposed fusion method has fewer such errors. The proposed fusion method successfully detects the inundation areas along the rivers, which could not be detected by the simple remote sensing image analysis. Even if the flood area is missing in the change detection results, our method succeeded in completing physically semantic disaster information in the missing area by using the change detection results of the surrounding area, such as debris flows in a mountainous area. Our approach failed to detect pixels having small absolute values and spatially small shapes in the reference simulation. When the changes are not detected by remote sensing image analysis due to occlusion by the tree canopy, it is difficult to estimate target variables that have very small patterns, such as a small/narrow debris flow, because each change (e.g., debris flow) occurs independently.

Fig.~\ref{hiroshima} shows the visual results of topographic deformation estimation for the Western Japan 2018 experiment. The advantage of our fusion method is much more evident. Both the reference simulation and the average simulation overestimate topographic deformation, while our result best resembles the LiDAR-derived reference. From Figs.~\ref{hiroshima}(c)(d), it can be observed that the estimated locations of debris flows are accurate and the overall scale of the disaster is similar, which was also confirmed by the quantitative evaluation in Table~\ref{tab:real}. We can find the inpainting capability in the green circles of Fig.~\ref{hiroshima}. Sediment deposition is not detected in remote sensing image analysis and is overestimated in the reference simulation. Our fusion method successfully estimates the sediment deposition using the hints of debris flow detection in upper streams. The estimation of sediment deposition is challenging as shown in the enlarged blue squares. The extent to which sediment is washed downstream depends largely on the disaster scenario. If the synthetic data do not include cases that are at least local but close to the real data scenario, our method may fail to estimate long-range displaced sediment deposition.

It should be noted that the reference simulation and the visual interpretation reference do not always match as shown in Figs.~\ref{realdataresults} and \ref{hiroshima}. This is evident numerically from the fact that the IoU of the reference simulation is much lower than 1 for both datasets. For example, the area indicated by the green circle in Fig.~\ref{realdataresults} is annotated as a debris flow area by human experts; however, the area is underestimated in the reference simulation. Since the proposed method uses the results of remote sensing image analysis as input and finds the corresponding topographical changes, our fusion method estimates that a large amount of sediments flowed out at places where debris flows were detected. This result is consistent with the visual interpretation reference, but does not match the simulation reference data, leading to higher IoU and RMSE, respectively.
Figs.~\ref{hiroshima}(a)(d)(e) clearly demonstrate the gap between the reference simulation and the visual interpretation reference and the LiDAR-derived reference for topographic deformation. The gap implies the difficulty of making a numerical simulation that matches the observation, even when the human-annotated debris flow starting points are used for the simulation due to the complex dynamics of mixed water and debris. Our framework overcomes this limitation by taking advantage of remote sensing, simulation, and deep learning.

\section{Conclusions and Future Lines of Enquiry}
In this paper, we proposed a framework that enables rapid estimation of the maximum water level and topographic deformation after simultaneous floods and debris flows through the use of remote sensing imagery and topographic data, a calculation that was not possible by using remote sensing image analysis and simulation alone. Our framework generates synthetic data of target variables and corresponding binary change maps based on simulation. It trains CNN-based regression models that take a binary change map and DEM as input and produce the maximum water level and topographic deformation as output. The CNN-based regression model can compensate for the missing part of the input detection map, which simplifies change detection and makes the whole process automatic and fast. Experiments based on two disaster events demonstrated the effectiveness of our framework both quantitatively and qualitatively.

In our future research, we intend to develop techniques that allow us to directly use remote sensing images instead of change maps as input for regression as well as techniques to scale up the system so that it can function over a much larger area.

\section*{Acknowledgment}
The authors would like to thank Hiroshima Prefecture for providing the LiDAR data used in this work.

\ifCLASSOPTIONcaptionsoff
  \newpage
\fi



%\bibliographystyle{IEEEtran}
\bibliographystyle{IEEEtran}
\bibliography{references,yamanoi}
\end{document}